# Algorithms For Automatic Accentuation And Transcription Of Russian Texts In Speech Recognition Systems

Yakovenko O.S., Bondarenko I.Yu., Borovikova M.N., Vodolazsky D.I.

Novosibirsk State University, Pirogova 1, 630090 Novosibirsk, Russia

```
olya.yakovenko@bk.ru
i.yu.bondarenko@gmail.com
m.borovikova@g.nsu.ru
daniil.vodolazsky@mail.ru
```

**Abstract.** This paper presents an overview of rule-based system for automatic accentuation and phonemic transcription of Russian texts for speech connected tasks, such as Automatic Speech Recognition (ASR). Two parts of the developed system, accentuation and transcription, use different approaches to achieve correct phonemic representations of input phrases. Accentuation is based on "Grammatical dictionary of the Russian language" of A.A. Zaliznyak and wiktionary corpus. To distinguish homographs, the accentuation system also utilises morphological information of the sentences based on Recurrent Neural Networks (RNN). Transcription algorithms apply the rules presented in the monograph of B.M. Lobanov and L.I. Tsirulnik "Computer Synthesis and Voice Cloning". The rules described in the present paper are implemented in an open-source module, which can be of use to any scientific study connected to ASR or Speech To Text (STT) tasks. Automatically marked up text annotations of the Russian Voxforge database were used as training data for an acoustic model in CMU Sphinx. The resulting acoustic model was evaluated on cross-validation, mean Word Accuracy being 71.2%. The developed toolkit is written in the Python language and is accessible on GitHub for any researcher interested.

## 1  Introduction

The main difficulty in Natural Language Processing (NLP) tasks is the irregularity of natural language in comparison to artificial language. This irregularity could be met at all levels, including the phonetic level. Moreover, the ambiguity of the phonetic representations in Russian written texts is a lot stronger than that in many other alphabetical languages. This is why, building an automatic speech synthesis or recognition system, it is important to introduce the most accurate phonemic representations of words and phrases.

Complex tasks such as building an acoustic model for ASR, apart from an effective inner architecture (Hidden Markov Models, Neural Networks), require first of all a big corpus with quality mark-up. As mentioned earlier, this quality can be achieved by correct phonemic representation of the spoken data which would serve as a foundation training set for an acoustic model. Modern open-source tools for automatic text accentuation and transcription are based mostly on letter representations of words and too little on possible phonetic features. Our approach is based on real phonetic process modeling in the spoken Russian language.

The goal of this paper is to introduce an approach for quick automatic accentuation and transcription of big texts based on rules, thesaurus and morphological information. We believe our research will be helpful for ASR or TTS projects which do not have the resources to do manual mark-up or build an enormous corpus but still require good results in recognition or synthesis of Russian language.

Although the problem of automatic text transcription is already thoroughly studied, with numerous complex algorithms introduced on a theoretical level, there are no quality grapheme-to-phoneme systems available in open-source. Thus, our goal is not solely to develop a system of automatic accentuation and transcription based on the research of Lobanov and Tsirulnik, but more importantly to fully develop the following system, reconsider the rules to correctly fit the realities of spoken Russian language and implement reconsidered algorithms in an open-source module, ready to be used by any researcher interested.

## 2     Related Works

Naturally, the investigations of the connection between phonemic and graphical representations of words, as of phonetic processes in general have a long history. Among the fundamental works dedicated to Russian phonetics there are studies of Avanesov (e.g. [1]), Vorontsova (et al. [2]), Zaliznyak [12]. The following works are used as reference to most automatised grapheme-to-phoneme systems, regarding both presented rules and examples.

One of the first studies, presenting a rule-based approach towards automatic transcription of Russian texts was authored by Krivnova et al. and published in 1970 [13]. Later, Krivnova et al. were working on a rule-based grapheme-to-phoneme system for TTS introduced in [7], which develops the idea of different levels of abstraction and discrimination of different allophones for specifying different types of sounds. Other rule-based transcription system is presented in [6], this paper features distinction of alternative transcriptions for words and phrases. The algorithm is implemented in a module written in C++ language.

Besides rule-based approaches, nowadays there is a scientific tendency for statistical analysis and Machine Learning (ML) to be used in NLP tasks including ASR and TTS. Such methods imply stress or phoneme prediction from the graphical representation of words [3,4,9]. Unfortunately, in application to the following NLP tasks, ML is most effective with big marked up data which researchers rarely have.

Disregarding the amount of research that already went into the problem of automatic transcription of Russian texts, there are only a few tools for automatic transcription of Russian texts available in open-source (e.g. epitran[1], Zamir Ostroukhov's tex-

---

1   https://github.com/dmort27/epitran

t2dict[2] or g2p_seq2seq[3] package based on machine learning). Otherwise, most quality grapheme-to-phoneme systems are parts of commercial products, thus remaining private to this date.

Our rule-based system of transcription is generally relied on Lobanov and Tsirulnik [8]. The main focus of this study is the modeling of phonetic processes in Russian language using general computerized assumptions about grapheme to phoneme mapping and basic phonetic processes between groups of phonemes (e.g. assimilation, reduction). The accentuation is founded on the electronic thesaurus[4] compiled from the grammatical dictionary of Zaliznyak [12].

The system presented in this paper is aimed first of all at such open-source products for ASR as Kaldi [10], CMU Sphinx [5] and HTK [11]. Those toolkits simplify and automatize the process of custom acoustic model production, based on both Hidden Markov Models and Neural Networks. Finally, they also feature a decoder for spoken data which uses pre-trained acoustic model to map the acoustic signals to phonemes.

## 3       Accentuation

To start with, the complexity of accentuation for the Russian language is rather high in comparison to some popular alphabetic languages (English, French). This is connected with a linguistic feature which was named "dynamic stress". It means that a certain graphical representation of a word does not have a fixed stress: e.g. *за́ ногу* or *за но́гу*, *о́кна* or *окна́*. This phenomenon illustrated by the examples (when the same graphical representation of the word might represent different pronunciations) is named homography. Thus, homography becomes the key issue behind automatic accentuation. The system described in the present paper has been developed in accordance with homograph resolution techniques based on word morphology.

As mentioned earlier, the developed system consists of two consecutive parts, the accentuation and the transcription. In this chapter the accentuation module is presented.

### 3.1       Main Corpus

The stress information of the most frequent Russian words has been collected from the electronic resource based on grammatical word analysis presented in [12]. The offline dictionary for the accentuation module was built with the help of the following resource. It includes a simple list of words with stress (which have an unambiguous stress variant) and a more complex data structure for homographs. Homographs in the dictionary are listed without stress and have different morphological information and the corresponding graphical representations with accent for the listed morphological forms. Morphological information is displayed in the Universal Dependencies[5] format and was generated initially by a Python module[6] based on RNN approach. The stress is marked as a "+" symbol after the stressed vowel; below in 1 is the sample from the dictionary described.

---

2   https://github.com/zamiron/ru4sphinx/tree/master/text2dict
3   https://github.com/cmusphinx/g2p-seq2seq
4   http://www.speakrus.ru/dict/#paradigma
5   http://universaldependencies.org/

**Table 1.** Sample from the dictionary of accents

| Homonyms | Non-homonyms |
|---|---|
| …, "альплагеря": {<br>    "NOUN Animacy=Inan\|Case=Gen\|Gender=Masc\|Number=Sing": "альпла+геря",<br>    "NOUN Animacy=Inan\|Case=Nom\|Gender=Masc\|Number=Plur": "альплагеря+"<br>    }, ... | ...<br>"проистека+л",<br>"проистека+ла",<br>"проистека+ли",<br>"проистека+ло",<br>... |

### 3.2　Architecture

The system waits for a tokenized sentence as input and also can receive morphological tags in the Universal Dependencies format, which could be generated by a Python module mentioned earlier for better homograph disambiguation. Each word of the phrase is processed separately; the steps for processing are the following:

1. Look up in the non-homonyms list;
2. Look up in the homonym list, choosing the exact form that corresponds to the word and the tag;

If the match could not be found in the offline corpus:

3. Search the exact word in wiktionary;
4. Save the result for future extraction and for rewriting the main offline dictionary.

The algorithm can also handle non-Russian words and symbols, anticipatorily accented words, function words and hyphenated compounds. Examples are in Table 2 below:

**Table 2.** Examples of automatic accentuation

| Input phrase | Morphotags | Result |
|---|---|---|
| ['подарок', 'для', 'кума'] | ['NOUN Animacy=Inan\|Case=Nom\|Gender=Masc\|Number=Sing', 'ADP _', 'NOUN Animacy=Anim\|Case=Gen\|Gender=Masc\|Number=Sing'] | ['пода+рок', 'для', 'ку+ма'] |
| ['фёдор', 'любит', 'кофе'] | - | ['фё+дор', 'лю+бит', 'ко+фе'] |

The quality of automatic phonemic transcription for ASR and TTS tasks depends highly on the stress placement in the word. This is the reason why automatic accentuation is an important part of the transcription.

---

6　https://github.com/IlyaGusev/rnnmorph

## 4      Transcription

A phoneme is one of the units of sound that distinguish one word from another in a particular language. Briefly speaking, if different words can be distinguished by a substitution of a phone, then that phone is a phoneme (e.g. free /fɹiː/ and three /θɹiː/). Like in many alphabetical languages with a big history, Russian graphics does not represent fully the phoneme composition of the word; for that cause simple "grapheme to acoustic signal" mapping shows good results only in the cases of huge data amounts. Smaller data require quality mark-up, and it could be achieved with rule-based transcription. As mentioned earlier, the transcription algorithm is based on Lobanov and Tsirulnik rules of phonetic transformations (pp. 35-37, 148 in [8]). Those rules correctly illustrate the correlation between graphemes and phonemes in the modern Russian language and model some of the most frequent phonetic transformations in the Russian spoken discourse.

### 4.1      Phoneme Set

There are several approaches to Russian system phonemes' definition. The most common are the following. Moscow phonological school (MPS) considers that there are 42 phonemes: 6 vowels and 36 consonants. In Leningrad phonological school (LPS) phoneme Ы isn't taken into account. We used the concept of phoneme as such from LPS but we are mindful for the Ы according to MPS. This combination was chosen due to the necessity of specifying most probable and significant acoustic events, since the algorithm is first of all aimed at creating transcriptions for manual speech recognition system training. On the one hand, it is important that this system recognize the Ы phoneme but on the other hand, the LPS's understanding of the phoneme definition is more efficient for such purpose given that there are many errors connected with aphetic phones. For ASR and even more for TTS it is important to distinguish stressed and non-stressed vowels in the word, since the spectral characteristics of the vowels differ greatly depending on stress. Therefore, a new set of "phonemes" has been designed to better illustrate the differences between stressed and non-stressed allophones of a phoneme (stressed are marked by zeros "0") and allophones which appear only in a certain phonetic context. A detailed description of the phoneme set could be found in our GitHub russian_g2p repository[7].

One other designation that the following system operates with is the "sil" quazi-phoneme that represents the absence of any speech data in the transcription: pauses, endings or beginnings of the recording.

The system waits for a tokenized sentence as input and also can receive morphological tags in the Universal Dependencies format, which could be generated by a Python module mentioned earlier for better homograph disambiguation. Each word of the phrase is processed separately; the steps for processing are the following:

### 4.2      Letter-phoneme transformation rules (TLP)

**Rule TLP 0.** If the whole word is in exclusion dictionary, it is converted in a form that will be transformed correctly by following rules. The exclusion dictionary has the following format:

---

7  https://github.com/nsu-ai/russian_g2p/blob/master/phoneme_description.pdfTranscription

автоби+знесу автоби+знэсу
оттого+ оттово+
здра+вствуй здра+ствуй

This way the algorithm preserves the stictness of the rules, also making it easy to add new exclusions for different contributors.

**Preliminary transformation rules.**

*Rule TLP 1.* In combinations of letters «СТН», «СТЛ» and «НТГ» letter Т is removed.

*Rule TLP 2.* In combinations of letters «ЗДН», «ЗДЦ», «НДЦ», «РДЦ», «НДШ» and «ГДТ» letter Д is removed.

*Rule TLP 3.* In combinations of letters «ЛНЦ» letter Л is removed.

*Rule TLP 4.* Combinations of letters «СЧ» и «ЖЧ» go to «Щ», «ТС», «ТЬС», «ТЦ», «ДС» and «ДЦ» go to «Ц», «СШ» goes to «Ш», «ЗЖ» goes to «Ж».

*Rule TLP 5.* Affixes «ОГО», «ЕГО» or «ГО» go to «ВО», except «много», «дорого», «строго» and some other adverbs.

**Main transformation module.** Every rule in this module are applied consequently to every letter in the phonetic syntagm from the last to the first (from right to left).

*Rule TLP 6.*
Modern mode:
  If i is «Н» and i+1 is one of the [J0], [TSH0], [SH0], [DZH0], [ZH0], [D0], [T0], [Z0], [S0], «Н» goes to [N].
  If i is «С», «З» and i+1 is one of the:
    1. [D0], [Z0] then i transforms to palatalized and voiced;
    2. [T0], [S0] then i transforms to palatalized and unvoiced;
    3. [N0]then i transforms to palatalized.
Classic mode:
  If i is «Н» and i+1 is one of the [J0], [TSH0], [SH0], [DZH0], [ZH0], [D0], [T0], [Z0], [S0], [L0], [M0], [P0], [B0], [V0], [F0], [N0], «Н» goes to [N].
  If i is «Т», «С», «Д», «З», «П», «Б», «В», «Ф» and i+1 is one of the:
    1. [D0], [Z0], [B0] then i transforms to palatalized and voiced;
    2. [T0], [S0], [P0] then i transforms to palatalized and unvoiced;
    3. [N0], [L0], [M0], [V0], [F0] then i transforms to palatalized.

This rule allows to make selective grapheme-to-phoneme transformations, unique for a certain dialect. For now the following module supports modern and classic mode, but it is planned to implement many more.

*Rule TLP 7.* Vowel transformation. Every vowel is being transformed consequently from right to left dependent on its context (previous and next letter). The transformation patterns are introduced in Table 3.

*Rule TLP 7.* Consonant transformation. Every consonant is being transformed consequently from right to left dependent on its context (previous and next letter). The transformation patterns are introduced in the table, available at GitHub russian_g2p repository[8].

**Postprocessing.**

---

8  https://github.com/nsu-ai/russian_g2p/blob/master/consonant_transformation_description.pdf

*Rule TLP 9.* If i and i+1 are the same phonemes, then they are substituted with a corresponding long variant (e.g. [A0 A0] → [A0l]).
*Rule TLP 10.* If i is a prefix of i+1, then they are substituted with a corresponding long variant of the i+1 phoneme (e.g. [Z ZH] → [Zhl]).

**Table 3.** Table captions should be placed above the tables.

| Position | | Stressed (+0) | | | | | | | | | Not stressed | | | | | | | | |
|---|---|---|---|---|---|---|---|---|---|---|---|---|---|---|---|---|---|---|---|
| i-1 | i+1 | Ё | Ю | Я | Е | О | У | А | Э | Ы | И | Ё | Ю | Я | Е | О | У | А | Э | Ы | И |
| Ъ Ь | nl[9] | JO | JU | JA | JE | JO | U | A | E | Y | JI | JO | JU | JA | JI | JA | U | A | Y | Y | JI |
| nl | nl[10] | JO | JU | JA | JE | O | U | A | E | Y | I | JO | JU | JA | JI | A | U | A | Y | Y | I |
| v | nl | JO | JU | JA | JE | O | U | A | E | Y | I | JO | JU | JA | JI | A | U | A | Y | Y | I |
| Ъ Ь | l | JO | JU | JA | JE | JO | U | A | E | Y | JI | JO | JU | JI | JI | JA | U | A | Y | Y | JI |
| nl | l | JO | JU | JA | JE | O | U | A | E | Y | I | JO | JU | JI | JI | A | U | A | Y | Y | I |
| v[11] | l | JO | JU | JA | JE | O | U | A | E | Y | I | JO | JU | JI | JI | A | U | A | Y | Y | I |
| Й Ч Щ | nl | O | U | A | E | O | U | A | E | Y | I | O | U | A | I | A | U | A | Y | Y | I |
| Й Ч Щ | l | O | U | A | E | O | U | A | E | Y | I | O | U | I | I | I | U | I | Y | Y | I |
| Ж Ц Ш | nl | O | U | A | E | O | U | A | E | Y | Y | O | U | A | Y | A | U | A | Y | Y | Y |
| Ж Ц Ш | l | O | U | A | E | O | U | A | E | Y | Y | O | U | Y | Y | A | U | A | Y | Y | Y |
| ps/ph[12] | nl | O | U | A | E | O | U | A | E | Y | I | O | U | A | I | A | U | A | Y | Y | I |
| ps/ph | l | O | U | A | E | O | U | A | E | Y | I | O | U | I | I | A | U | A | Y | Y | I |

## 5      Evaluation of the resulting acoustic model

Several experiments were conducted in order to test the quality of the generated transcriptions. CMU Sphinx was used as an interface to training an acoustic model and decoding utterances using the acoustic models that were previously used.

### 5.1      Corpus

Training and testing was carried out on a set of utterances, available at [14]. Corpus of Russian speech consisted of an audio set and a set of text annotations that describe the spoken phrase. Text annotations were preprocessed, inserting <sil> quaziphonemes in place of punctuation and creating phonetic dictionaries based on the unique vocabulary of the train set. Resulting corpus consisted of 8831 different utterances including 646 unique speakers, summing up to a total length of 20 hours. The audio was recorded in the conditions of minor noise, with the sampling rate of 8 kHz, bit depth of 8 bit and written in a mono channel.

---

9   not a letter
10  letter
11  vowel
12  paired consonants (voiced and unvoiced)

## 5.2 Experiments

Three open-source grapheme-to-phoneme systems that were mentioned earlier are tested on the presented dataset. Those systems are: russian_g2p (presented in the following paper), text2dict and epitran.

For all of the experiments the same set of standard parameters was used. An acoustic signal undergoes a discrete cosine transformation (DCT) with the lowest frequency limit set to 200 Hz and highest – to 3500 Hz. 13 MFCCs were used in the acoustic front end and fed to a HMM with a total number of 3000 senones. Total of 3 acoustic models were built using the different phonetic dictionaries generated by three different grapheme-to-phoneme algorithms.

For the evaluation of the resulting acoustic models decoding was performed for both word recognition using n-gram language modeling and phoneme recognition separately. Like earlier, the outcome will be presented in the terms of WER and PER. The results of the experiments were cross-validated on the total of 10 folds, a mean for each decoding iteration is shown below in Table 4 below:

**Table 4.** Examples of automatic accentuation

|     | russian_g2p | text2dict | epitran |
|-----|-------------|-----------|---------|
| WER | **28,80%**  | **27,88%** | **29,80%** |
| PER | **31,89%**  | **32,19%** | **34,36%** |

## 6 Conclusion

The present study shows that the developed system is capable of fast and accurate accentuation as well as transcription of Russian texts. The accent dictionaries complemented by wiktionary parsing and morphological information cover many cases of graphical ambiguity. Generated transcriptions reflect many processes in Russian spoken discourse such as vowel reduction, assimilation and simplification of consonant groups. These grapheme-to-phoneme transformations do not only work within a single word but also at the junctions of words. This correctly reflects natural spoken language processes, as the ability to implement different dialects can fully represent the variability among native speakers of the Russian language. This results in the system showing decent quality of automatic phonemic mark up of isolated words in speech recognition tasks. The resulting acoustic model does not concede to other open-source systems, while the system's potential in automatic transcription for ASR tasks is yet to be unfolded. The advantage of the system consists of precise transcriptions of the whole phrases, but not isolated words. This way the next step should be making use of one of the main features of the developed algorithm – that is training a model with fully transcribed sentences and, therefore, building a better acoustic model.

Naturally, there are several things to improve in the algorithm itself, for example:

- Implementation of Wikipedia parsing into the accentuation stage that will lead to the expansion of the dictionary with proper names;
- Machine learning based supplementation for accenting unknown words or handling remaining word ambiguity;
- Implementation of many more dialects;
- Extension of the transcription rule set towards more precise and complex cases (such as affix handling in adjectives).

Finally, it is necessary to mention that this approach has been implemented in a module, written in Python language[13] and is ready for use by any researcher interested. We believe that the present study will boost interest towards open-source development of ASR and TTS systems and hope that it will be of use to the scientific community.

## Acknowledgements

- This paper was supported by Botan Investments[14].
- Special thanks to Zamir Ostroukhov[15] for providing the script in open-source and being able to use the algorithm for evaluation.

13 https://github.com/nsu-ai/russian_g2p
14 http://botaninvestments.com/ml-program/
15 https://github.com/zamiron